\documentclass[10pt, letterpaper]{article}
\usepackage{cite}
\usepackage[hyphens]{url}
\usepackage{graphicx}
\usepackage{amsmath, amssymb}
\usepackage{lettrine}
\begin{document}

\title{\textbf{Open-Ended Wargames with Large Language Models}}
\author{Daniel P. Hogan \and Andrea Brennen}
\date{IQT Labs\\ \small{\url{{dhogan, abrennen}@iqt.org}}}
\maketitle

\begin{abstract}
Wargames are a powerful tool for understanding and rehearsing real-world decision making.  Automated play of wargames using artificial intelligence (AI) enables possibilities beyond those of human-conducted games, such as playing the game many times over to see a range of possible outcomes.  There are two categories of wargames: quantitative games, with discrete types of moves, and qualitative games, which revolve around open-ended responses.  Historically, automation efforts have focused on quantitative games, but large language models (LLMs) make it possible to automate qualitative wargames.  We introduce ``Snow Globe,'' an LLM-powered multi-agent system for playing qualitative wargames.  With Snow Globe, every stage of a text-based qualitative wargame from scenario preparation to post-game analysis can be optionally carried out by AI, humans, or a combination thereof.  We describe its software architecture conceptually and release an open-source implementation alongside this publication.  As case studies, we simulate a tabletop exercise about an AI incident response and a political wargame about a geopolitical crisis.  We discuss potential applications of the approach and how it fits into the broader wargaming ecosystem.
\end{abstract}

\section{Introduction}

Wargames can be an effective problem-solving tool to understand and address real-world challenges\cite{burns15, csis23}.  The value of using artificial intelligence (AI) for automated or semi-automated playing of wargames has long been recognized\cite{davis84, goodman20, davis22}.  However, previous automation efforts have largely ignored open-ended wargames in which player moves are not restricted to predefined options.  In this work, we show that large language models (LLMs) enable automation of such games.  We also introduce an example implementation called ``Snow Globe.''

\subsection{Quantitative and Qualitative Wargames}

We define a ``wargame'' as a reality-based decision-making game for one or more human players.  Following \cite{burns15, davis22}, we do not restrict the term to games that feature multiple adversaries as opposed to just one.  As a result, quoting \cite{davis22}, the term will be understood ``to include everything from small seminar exercises\ldots to large multi-day, multi-team wargames.''  Furthermore, a wargame does not have to be about war\cite{lingreenberg22}.  One can wargame a political crisis\cite{goldhamer59, pauly18}, a disaster response\cite{dhs20, dot14}, a technology problem\cite{cisa21, haiwire23}, or any other decision-making scenario.

Wargames are often subdivided based on their intended purpose.  In one common scheme\cite{dcdc17, culkin99, burns15, csis23}, a distinction is drawn between analytical / experimental wargames to study strategic and situational variations, versus educational / experiential wargames to train the players in decision-making.  An alternate approach \cite{bartels20, geist24} divides wargames into four categories (systems exploration, alternative conditions, innovation, evaluation) based on two binary considerations: understanding vs. addressing a problem, and early vs. mature research.

For the purpose of automating a wargame, however, the most important distinction is not why the game is played, but rather how it is played.  In terms of game mechanics, there are two broad categories of wargame.  In a ``quantitative'' wargame, there are discrete options for the kinds of things each player can do on their turn, and the rules of the game dictate the consequences of those options.  A ``qualitative'' wargame, on the other hand, is much more open-ended.  Each player on their turn verbally describes what they would do in that situation, and anything possible in real life is a valid response.  If the flow of the game depends on previous player decisions, the moderator adjudicates the outcome for that purpose.  For example, most tabletop exercises (also known as seminar games) are qualitative games, as are the political-military wargames that grew out of the work of the RAND Corporation's Social Sciences Division (SSD) during the Cold War\cite{emery21}.

The philosophical difference between a quantitative and a qualitative approach is most clearly articulated in \cite{emery21}, which takes as a prototypical example the dueling versions of a ``Cold War Game'' developed by RAND's Mathematical Analytics Division and the aforementioned SSD in the 1950s.  This dichotomy has also been described in the literature in different language: ``rigid games'' versus ``free play'' games\cite{lingreenberg22}, for example.

\subsection{Automated Wargames}

There are benefits to having humans play wargames, with the chief benefit being the educational value for the humans involved.  However, there are also benefits to being able to automate the playing of wargames.  One such benefit is the ability to play the game many times over, elucidating a full spectrum of possible outcomes.  Another benefit of automation is the ability to help humans play the game better.  For example, an AI could play along with human players, offering suggestions along the way.  Such considerations motivate the development of automated wargames, in which an automated system makes the moves instead of human players.  (Note that a fully automated wargame should technically be described as a ``simulated wargame,'' as it's not a true wargame in the absence of human players, a semantic distinction with which we will not be overly concerned in this text.)

For quantitative wargames, there is an extensive body of work about automating these types of games.  To begin, very simple quantitative wargames may be especially amenable to analysis with game theory.  This approach can sometimes shed light on counterintuitive aspects of effective strategies\cite{axelrod81, bravetti18}.  Game theoretic models have been developed for many aspects of international relations and military activity, although the practical usefulness of such effort has been the subject of debate\cite{oneill94}.

In more realistic and hence more complicated quantitative wargames, automation can be done via computer simulation of the game and its players.  A notable early example of a wargame simulation was the RAND Strategy Assessment System (RSAS) in the 1980s, which accommodated human players as well as AI players implemented as expert systems\cite{davis84, davis22}.  Using AI players to create wargame simulations has continued to the present day\cite{knack23}.  In recent years, reinforcement learning has been an effective technique for implementing such systems\cite{goodman20}.

Given that game theory and the existing simulation literature provide many tools for automating quantitative wargames, what is the equivalent approach for qualitative wargames?  In other words, how can we automate an open-ended wargame?  This problem has gone largely unaddressed in the literature because until recently it was intractable.  A solution requires generating textual output based on reasoning from textual input in the absence of task-specific training data.  Such a capability has only recently become available with the introduction of large language models\cite{vaswani17, bubeck23, huang23}.  In the next section, we describe an approach to leveraging LLMs to play qualitative wargames.  To keep the scope of this effort reasonable, we limit our focus to simulation of qualitative wargames that are entirely text-based.  This circumscribed problem space nevertheless spans a wide range of qualitative wargames, ranging from structurally simple games (like \cite{millot93}) to others that are quite complex (like \cite{saga72}).  However, this scope does exclude games where exact positions on a complicated game map play an important role (like \cite{mouat16}).  Given ongoing advances in multimodal LLMs\cite{yang23}, it may not be long before future work can step beyond this limit.

Finally, it should be noted that there have been several previous investigations that use LLMs to play wargames.  However, most of these have been built around quantitative games.  One effort\cite{bakhtin22} uses an LLM to play the board game ``Diplomacy.''  Although Diplomacy allows unconstrained player negotiation, actual player moves are constrained to those specified by the game's rules (attack, support, etc.).  Other efforts at LLM-driven wargame simulations\cite{hua23, rivera24} are similarly constrained, with the games in the cited works permitting exactly 7 and 27 discrete types of moves, respectively.  In another recent work, the LLM chooses from menus of prewritten options, for a total of 21 binary choices\cite{lamparth24}.

Perhaps the only prior work on qualitative wargames with LLMs concerns so-called ``matrix games'' (confusingly, not related to game theory matrix games) where adjudication is handled by listing reasons why a goal would or would not be achieved\cite{griffin23, brynen24}.  Additionally, some approaches using LLMs to play role-playing games and other games\cite{gallotta24} are transferable.

\begin{figure}[t]
\includegraphics[width=\linewidth]{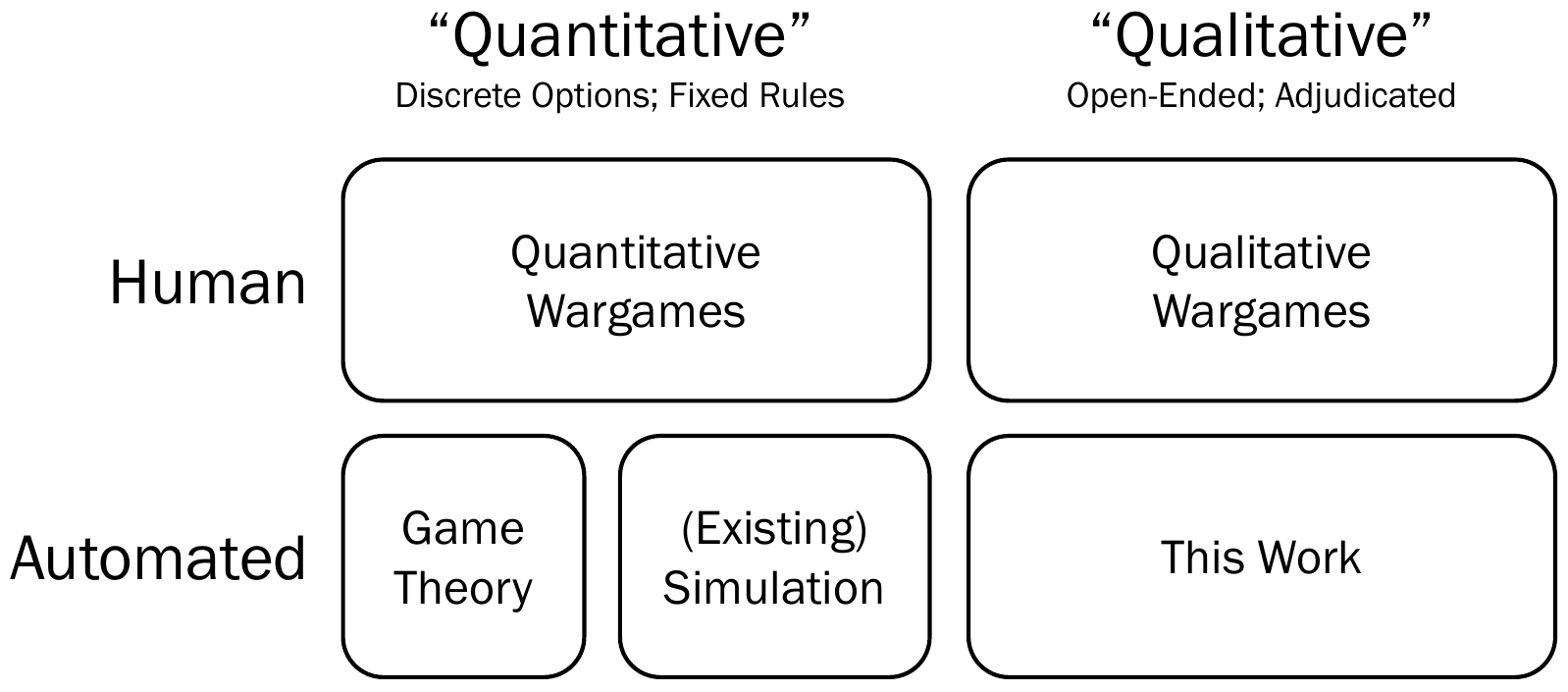}
\caption{Taxonomy of human and automated wargames.}
\label{fig_taxonomy}
\end{figure}

This discussion is summarized schematically in Figure \ref{fig_taxonomy}.

\section{Software Architecture}

Snow Globe is a multi-agent system designed to automate a wide range of qualitative wargames.\footnote{It is so named because a snow globe is a simulated snowstorm self-contained in a glass orb, and by analogy this work is a simulated crisis self-contained in software.}  In Snow Globe, there are three types of agents: control, players, and teams.  Of these, the control and players are LLM-powered intelligent agents which carry out the traditional human roles of a wargame.  The control is the game's moderator.  Its job includes purely logistical tasks, like calling the teams/players to get their responses, as well as complex AI-driven tasks, like adjudicating each round of an adversarial game via LLM.  Player agents simulate the responses of humans playing the game, and each can be optionally given a ``persona,'' a written description of their perspective that influences their responses.  Finally, a team agent simulates a group that must jointly formulate a response.  A simulation generally has only one control but can have any number of players and teams.  Any acyclic arrangement of teams/players is allowed, including a team of teams, a player who's on multiple teams, etc.  This is illustrated in Figure \ref{fig_architecture}.

\subsection{Conceptual Approach}

In Snow Globe, the record of what's happened so far in a wargame simulation is recorded in a ``history'' object.  A history object is an ordered list of text entries, each associated with an agent.  For games without information asymmetry, there is a common history maintained by the control.  Information asymmetry can be introduced by giving each player/team their own possibly incomplete history object.  The game record contained in a history object provides the context in which players make their responses.  Formally, the history object(s) constitute the environment of the multi-agent system.

Looking more closely at the different types of agents, the main task of players is to state how they would respond to each situation presented to them.  The prompt sent to a player's underlying LLM for this purpose includes the history described above, the written description of the player's persona, and the query, ``What action or actions do you take in response?''  (Specific questions can also be substituted for that general-purpose one.)  The players' personas affect what they emphasize in their answers.  For example, we've developed an information technology tabletop exercise simulation that includes players with ``data scientist'' and ``lawyer'' personas.  The former is more likely than the latter to raise concerns about training data, while the latter is more likely to mention legal compliance.

\begin{figure}[t]
\includegraphics[width=\linewidth]{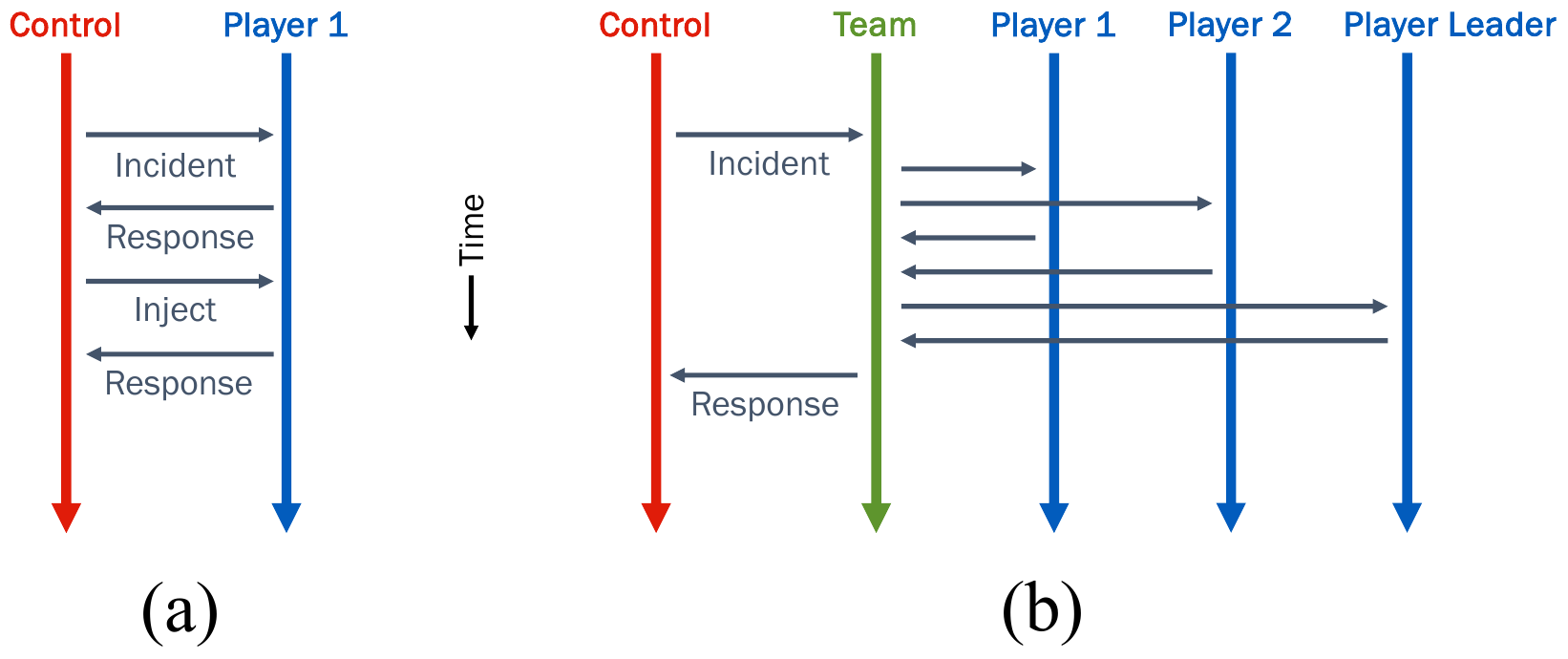}
\caption{Software architecture examples for the Snow Globe multi-agent system.  Vertical arrows represent agents and horizontal arrows represent information flow for (a) a simple tabletop exercise and (b) a team of players formulating a collective response.}
\label{fig_architecture}
\end{figure}

A team, from the control's point of view, looks like just a player -- it receives the current history and sends back a response.  However, a team has no generative capacity of its own.  Behind the scenes, the team asks each team member for their response, then asks a designated team leader player to combine those answers into a single joint response that's sent back on behalf of the team.  Figure \ref{fig_architecture}(b) shows this schematically.

The control, given its job as the game's moderator, has many tasks.  One control task is to adjudicate adversarial wargames.  For this, the control calls each player to find out what actions they propose in response to the current situation.  The control then generates a plausible narrative of the outcome, describing what could reasonably happen as the players' stated plans are carried out.  Adjudication is further discussed in the next paragraph.  The control can also help create the wargame by generating a situation based on a short prompt, identifying key players in that situation (for which player agents can be created), and generating relevant ``injects'' (pre-written plot developments used to drive the game).  Finally, the control provides tools to analyze the game as it is played or after it's completed.  The control can answer preset questions about the game, or a user can converse with it through a chatbot dialog.  In either case, the control draws on the game's history to answer questions.  Questions can range from specific (e.g., ``Was a lawyer consulted at any point?'') to general (e.g., ``What lesson can we learn from this scenario?'').

Adjudication is perhaps the most challenging task when implementing qualitative wargames with LLMs.  It requires the LLM to create a narrative with a logical sense of cause and effect, which may necessitate developing a plausible resolution when players are acting at cross-purposes.  We use the query, ``Weave these plans into a cohesive narrative of what happens in the next month,'' modifying the time step as appropriate.  In a realistic game, not everything important that happens is the result of a player setting out to make it happen.  For example, in a political-military wargame, there can be fast-moving developments on the ground that no leader ordered yet are central to the outcome (civil disturbance, terrorism, etc.).  To be useful, AI adjudication must be able to generate ``nature,'' as these contextually plausible outside developments are called\cite{emery21}.  This difficult task turns out to be surprisingly easy with LLMs.  All that's required is to modify the prompt by adding a three-word sentence: ``Include unexpected consequences.''

When adjudication does give low-quality results, the most common problems include repeated output from move to move and output that deviates from players' stated plans (usually by hallucinating new player plans as part of the output).  However, a hypothetical hallucination-free LLM would create more problems than it solves.  Hallucination, which in other applications is so often harmful, is the key to making open-ended LLM wargames work.  The well-grounded creativity required to generate plans and adjudicate their outcomes is hallucination by another name.

\subsection{Software Implementation}

We have developed a Python implementation of the conceptual architecture described above.  The Snow Globe software has been open-sourced\footnote{Available at \url{https://github.com/IQTLabs/snowglobe}} under the Apache 2.0 license.

After importing the Python module, users can create control, team, and player agents, as well as history and LLM objects.  A recommended approach is to create a child class of the module's control class, to serve as the control for the user's specific simulation.  The constructor and call methods of that child class can then be overridden with code to create the other agents and run the simulation, respectively.

The software supports locally run LLMs formatted for either llama.cpp\cite{gerganov24} or the Hugging Face transformers library\cite{huggingface24}.  It also supports OpenAI API calls\cite{openai20}.  The LangChain framework\cite{chase24} handles interactions with LLMs.  Among small, open-source LLMs, we've found the best results with the Mistral-7B-OpenOrca-GGUF model.  A seven-billion-parameter foundational model from Mistral AI\cite{mistral23} was fine-tuned as a chatbot using the OpenOrca dataset\cite{openorca23} and then converted to a computational resource-efficient format\cite{jobbins23} by other open-source contributors.  This LLM is lightweight enough to run on an ordinary ``gaming'' laptop.  The ability to run the model locally is crucial for use cases where information cannot be shared externally.  Because the model is not fine-tuned for any particular game (nor is it given example output at runtime), this is a case of zero-shot learning.

The software is configured so that LLM output is randomized, with the result that no two runs of a simulated wargame produce identical results.  Because the outcome of any complicated real-world situation hinges on many small details, not all of them characterized and furnished in a simulation, it is not reasonable to expect the simulation to produce a single ``correct'' answer.  (This is a characteristic of qualitative wargames generally that our system implementation mimics.)  Instead, repeated iterations of the same simulation with randomized LLM output show a range of possible outcomes.  An example of this is given below in Section \ref{geocrisis}.  The randomness also means that no agent knows exactly what another agent will do in a given situation.

\subsection{Human-Machine Interaction}

Snow Globe makes it possible for any player agent to be operated by an actual human instead of an LLM.  A human player uses a computer connected to the network where the simulation is running, and a web-based user interface tells the player the state of the game by displaying the prompts that otherwise would have been sent to the LLM.  The web page receives the human's written response to each prompt and sends it back to the server.

Any combination of AI and human players is supported, which permits a range of wargame possibilities.  For example, a human player could face an AI adversary.  Alternately, a human player could have a staff of AI assistants, each offering their input about what they think the human should do.  (The latter case could be implemented using a team agent with AI team members and a human team leader.  The assistants could all have different personas or even use different LLMs.)  It's also possible to have a human moderator by setting the control agent to use the web-based user interface.

When multiple players must respond to the same situation (each without knowing the others' responses), an issue of timing arises.  AI players should generate their responses sequentially to avoid overburdening the hardware.  Human players, however, should be able to work on their answers in parallel to avoid unnecessary waiting.  In short, AI is compute-bound and humans are I/O bound.  We use asynchronous programming to optimize both without requiring code refactoring to switch between AI and human players.

In our Python implementation of Snow Globe, prompts being sent to humans and responses being received from them are stored as text files, and an API with a FastAPI framework and Uvicorn web server makes them available over the network.  This future-proofs the interface by making it easily extensible.  One could design a new web page to communicate with the existing API or forgo the API and design new software to interact with the files directly, with no required changes to the underlying simulation code in either case.

Per \cite{bartels20}, there are four basic tasks associated with wargaming: preparing the game, playing the game, adjudicating the game, and interpreting the game.  For qualitative text-based wargames with Snow Globe, each one of these tasks can be optionally carried out by AI, humans, or a combination thereof.

\section{Example Simulations}

In this section, we demonstrate two very different simulated wargames: a tabletop exercise about an AI incident response and an adversarial wargame about a geopolitical crisis.

\subsection{Tabletop Exercise}\label{haiwire}

``HAIWIRE'' is an open-source tabletop exercise in which participants must formulate a response to an incident or problem involving the use of AI\cite{haiwire23}.  Participants draw a random incident then a series of random injects, furnishing a response for each one.  In our simulation of this tabletop exercise, a single AI player serves as a stand-in for the collective wisdom of these human participants, shown schematically in Figure \ref{fig_architecture}(a).

The beginning of an example game is shown in Table \ref{tab_haiwire_demo} in the appendix.  As can be seen, the AI player's response doesn't just address the issue on technical grounds but also considers the human element and institutional element of the situation to generate a well-rounded answer.  A full game may take a small group of humans twenty minutes or more; the simulation runs in less than one minute on a consumer GPU.

\subsection{Geopolitical Crisis Simulation}\label{geocrisis}

To demonstrate another kind of qualitative game, we introduce a political wargame.  The game scenario is a geopolitical crisis involving the fictional Central Asian countries of ``Azuristan'' and ``Crimsonia,'' both seeking to control a disputed province.  The game's AI players are the leaders of the two countries.  In each move of the game, both players state their plans (without knowing the other player's plans), and then the result is adjudicated by the control.  The number of moves is set in advance.  Table \ref{tab_geocrisis_demo} in the appendix shows the scenario and the first move of an example game.

One test of the system's realism is whether the personas of the players affect the outcome of the game.  To explore this, we consider two possible personas for each player: a ``dove'' persona where they prioritize peace above all else, and a ``hawk'' persona where they prioritize other national interests.  These are shown in Table \ref{tab_personas}.  For each of the three combinations of personas shown in Table \ref{tab_conflict}, the simulation is run 20 times with a single one-year move each.  Unsurprisingly, armed conflict is seen more often when more hawks are involved.  Although we would caution against placing too much credence in the raw probabilities, this case study illustrates that repeated iterations of a qualitative wargame can show causal relationships between policies and outcomes and even between personalities and outcomes.

That's significant because a previous LLM wargame study (with a quantitative game) did not see such a relationship, concluding that ``the LLMs are inadequate at accounting for player backgrounds\ldots'' \cite{lamparth24}.  Here, however, the effect is evident.

\begin{table}[t]
\centering
\begin{tabular}{|c||p{0.95in}|p{0.95in}|}
\hline
& Azuristan & Crimsonia \\
\hline
\hline
Dove
& Your goal is to avoid war at all costs, and to preserve the sovereignty of Azuristan if possible.
& Your goal is to avoid war at all costs, and to unify the Crimsonian people if possible.
\\
\hline
Hawk
& Your goal is to preserve the sovereignty of Azuristan, even if it requires starting a war.
& Your goal is to unify the Crimsonian people, even if it requires starting a war.
\\
\hline
\end{tabular}
\caption{Two possible personas for each of the two AI players in the geopolitical crisis simulation.}
\label{tab_personas}
\end{table}

\begin{table}[t]
\centering
\begin{tabular}{|c|c||c|}
\hline
Azuristan & Crimsonia & Armed Conflict \\
\hline
\hline
Dove & Dove & \hphantom{0}1 / 20 \\
\hline
Dove & Hawk & \hphantom{0}4 / 20 \\
\hline
Hawk & Hawk & 14 / 20 \\
\hline
\end{tabular}
\caption{Frequency of armed conflict in the geopolitical crisis simulation for each of three combinations of AI player personas.}
\label{tab_conflict}
\end{table}

\section{Conclusion}

Large language models make possible a new kind of simulated wargame.  With LLMs, qualitative wargames can be automated, something that was previously only possible for quantitative wargames.  AI, like humans, can now engage with open-ended games in which actions and outcomes are not \emph{a priori} limited by the game's design.

In this paper, we have discussed where automated qualitative wargames fit into the broader wargaming ecosystem.  We have described a flexible software architecture for LLM-powered text-based qualitative wargames.  In addition to a conceptual description, we also release a Python implementation.  Examples of a tabletop exercise and a political wargame demonstrate the method's utility.

With LLM-based wargame automation, the same software can be redirected to new problems with ease.  As a historical comparison, a progress report on the RSAS automated wargame predicted that building fewer than half a dozen escalation models (effectively, personas) from sources such as ``unstructured prose description'' would be ``both painful and interesting'' and would ``occupy [the authors] through most of 1984'' \cite{davis84}.  Contrast that with Snow Globe, where the prose description is furnished directly to the software for immediate use.  Although RSAS could explore its specific problem domain (``large-scale conflicts'') in much greater detail than Snow Globe, the latter can address challenges across many domains, even something as unrelated as a misfunctioning robot (Table \ref{tab_haiwire_demo}).  LLMs enable rapid adaptability and a wide scope of applications.

An equally important advantage of LLMs is allowing unconstrained moves from players, both AI and human, and unconstrained developments introduced during adjudication.  In theory, an ideal LLM-based qualitative wargame simulation is restricted only by what's plausible, not by what the simulation writer had the time (or creativity) to code up.

With further development, a system like Snow Globe could be directed towards many potential applications.  It could become a tool to help organizations solve problems, like in the HAIWIRE example.  It could become a platform for automated wargaming, like in the geopolitical crisis simulation example.  It could become a testbed for human-AI teaming.  There will always be a place for humans in wargaming.  Discovering the place for LLMs is only just beginning.

\bibliographystyle{ieeetr}
\bibliography{snowglobe}

\appendix
\section*{Appendix}

Table \ref{tab_haiwire_demo} shows the start of a HAIWIRE simulation (Section \ref{haiwire}), including the incident and first inject.  Table \ref{tab_geocrisis_demo} shows the start of a geopolitical crisis simulation (Section \ref{geocrisis}), including the scenario and first move.

\begin{table*}[p]
\makebox[\textwidth][c]{
\begin{tabular}{|p{1.3\textwidth}|}
\hline
\small
\emph{``King's Ransom'':} \\
\small
\emph{Your company's chess-playing robot fractured the finger of its 7-year-old opponent during a tournament.} \\
\hline
\footnotesize
As the management of a technology company, our primary concern would be to ensure the wellbeing and safety of all parties involved, especially the young competitor who was injured in the incident. Here are some actions we would take:

1. Immediate Apology and Support: We would issue an immediate apology to the young competitor and their family, expressing our deep regret and concern for their wellbeing. We would offer full support for their medical treatment and any additional care they might require.

2. Investigation and Analysis: We would conduct a thorough investigation into the incident, reviewing all aspects of the robot's design, programming, and operation to determine if there were any flaws or defects that contributed to the injury. We would also analyze video footage and consult with experts in robotics, engineering, and safety to gain further insight into the cause of the accident.

3. Product Recall and Upgrades: If we find any issues with our robot, we will initiate a product recall and implement necessary upgrades to ensure the safety and reliability of our product. We would also develop and implement strict safety protocols and guidelines for all future tournaments involving our robots, as well as regular safety inspections and testing.

4. Enhanced Safety Measures and Training: In addition to product recalls and upgrades, we will also enhance safety measures and training for all staff involved in operating and supervising our robots, including regular safety training sessions and drills.

5. Communication with Stakeholders: We will keep all relevant stakeholders, including the child's family, the chess community, and the public, informed of our actions, findings, and improvements, ensuring transparency and accountability throughout the process.

6. Long-term Prevention Measures: Lastly, we will continue to research and develop safer and more advanced robotic technologies, incorporating lessons learned from this incident to ensure that our future products prioritize safety and performance. \\
\hline
\small
\emph{``Need to Know Basis'':} \\
\small
\emph{Employees are infuriated about the lack of transparency within your organization about this mishap and subsequent actions taken.  Most of them learned about this incident either online or from outside contacts and are in the dark about what is going on to resolve.} \\
\hline
\footnotesize
As the management of a technology company, we understand that our employees play a crucial role in maintaining our organization's reputation and ensuring customer satisfaction. In light of the recent incident involving our chess-playing robot and concerns about transparency, we will take the following steps to address these concerns and improve communication within our organization:

1. Internal Communication and Transparency: We will establish regular, open forums for employees to discuss any concerns or questions they may have regarding company policies, procedures, and actions taken in response to incidents like this one. We will encourage an open exchange of ideas and ensure that all relevant information is shared with all employees.

2. External Communication and Transparency: We will work on improving our external communication channels to keep all relevant stakeholders, including employees, informed about our actions and findings related to this incident and any other incidents that may arise in the future. This will be achieved through regular press releases, social media updates, and an improved internal communication infrastructure.

3. Employee Involvement and Training: We will create opportunities for employees to participate in incident analysis and resolution efforts, helping to ensure that all perspectives are considered and that all necessary measures are taken to prevent future incidents. We will also provide training on effective communication and transparency best practices to help employees better navigate sensitive situations and maintain trust within our organization.

4. Continuous Improvement and Learning: We will continually evaluate our internal and external communication practices and policies, using feedback from employees and stakeholders to improve our overall communication and transparency efforts. We will also use this incident as an opportunity to learn, grow, and further prioritize safety and transparency throughout our organization. \\
\hline
\end{tabular}
}
\caption{Example output from a game of HAIWIRE, showing the incident and the first inject.  Italic text is prewritten, and roman text is LLM-generated.}
\label{tab_haiwire_demo}
\end{table*}

\begin{table*}[p]
\vspace*{-1in}
\makebox[\textwidth][c]{
\begin{tabular}{|p{1.5\textwidth}|}
\hline
\footnotesize
\emph{Azuristan and Crimsonia are neighboring countries in Central Asia.  Azuristan is a Western-backed democracy that suffers from endemic corruption.  Crimsonia is controlled by an autocratic government that stifles dissent and commits human rights violations.}
\emph{Both countries have modern professional militaries, although Crimsonia's is slightly larger than Azuristan's.  In addition, Azuristan possesses ten nuclear weapons and Crimsonia has eight.}
\emph{Most citizens of Azuristan are from the Azuristani ethnic group, and most citizens of Crimsonia are from the Crimsonian ethnic group.  The only exception is Azuristan's province of Tyriana, on the border with Crimsonia.  Most residents of Tyriana belong to the Crimsonian ethnic group.}
\emph{The animosity between Azuristan and Crimsonia extends back over centuries of ethnic tension and intermittent warfare.  Recent years have been fairly calm.  However, that suddenly changes when Tyriana declares independence.  Amidst the crisis, local leaders in Tyriana ask Crimsonia to come to the province's defense, and the same leaders indicate that they want Tyriana to become part of Crimsonia.} \\
\hline
\small
\emph{Move 1} \\
\hline
\small
\emph{President of Azuristan (With ``Dove'' Persona):} \\
\footnotesize
In response to Tyriana's declaration of independence and its request for assistance from Crimsonia, I would take the following actions:

1. Immediately initiate diplomatic talks with both Tyriana's leaders and the government of Crimsonia to try to resolve the situation peacefully. I would emphasize that Azuristan respects the right of Tyriana to self-determination but expects the province to remain a part of Azuristan.

2. Offer economic incentives to Tyriana for staying within Azuristan's borders. These could include investing in infrastructure projects, providing financial support for development initiatives, or offering tax breaks for businesses in Tyriana.

3. Strengthen the security presence in Azuristan's border with Crimsonia to prevent any potential military escalation between our two countries. This could involve deploying more troops along the border, increasing surveillance capabilities, and coordinating with our regional allies to ensure a strong defense against potential aggression from Crimsonia.

4. Engage in a public relations campaign aimed at promoting unity within Azuristan. This would involve highlighting our shared culture, history, and interests while downplaying ethnic differences. The aim would be to reduce tensions between different ethnic groups within our country and discourage any further separatist movements.

5. Maintain close contact with the international community, particularly with countries that have influence over both Azuristan and Crimsonia, such as Western allies and regional powers. These countries could potentially help mediate the situation or apply diplomatic pressure on Crimsonia to back down from its support for Tyriana's independence movement.

6. If all else fails and it appears that military conflict may be imminent, I would consider offering further concessions to Tyriana or even entering into negotiations for a peaceful division of Azuristan, in order to prevent an all-out war that could jeopardize both countries' stability and security. \\
\hline
\small
\emph{Premier of Crimsonia (With ``Dove'' Persona):} \\
\footnotesize
As the leader of Crimsonia, I will take the following actions to address this crisis:

1. Assess the situation: First, I will gather information on the situation in Tyriana, including the reasons for declaring independence and the level of support for this move among Tyrian citizens. I will also analyze how this could affect relations with Azuristan and any potential impact on regional stability.

2. Offer support and diplomacy: I will offer support to Tyria in their struggle for independence while simultaneously engaging in diplomatic efforts with Azuristan. I will propose a peaceful resolution that would allow Tyriana to maintain its autonomy while ensuring that regional stability is maintained. This could involve international mediation or negotiations between the two nations.

3. Strengthen the Crimsonian people's unity: In order to achieve our goal of unity among the Crimsonian people, I will emphasize our shared cultural heritage and history in my public speeches and messaging. I will also encourage dialogue between citizens of Tyriana and the rest of Crimsonia to promote understanding and mutual respect.

4. Maintain a strong but restrained military posture: While I will ensure that our military is prepared for any potential conflict, I will also be cautious in how we project our power and avoid escalating tensions with Azuristan. This may involve refraining from sending troops into Tyriana or engaging in aggressive military maneuvers near Azuristan's border.

5. Foster economic cooperation: To further promote peace between our nations, I will encourage trade and economic cooperation between Azuristan and Crimsonia. This could involve signing mutual trade agreements or forming joint ventures in various industries. By creating economic incentives for peace, we can make it more difficult for political tensions to lead to armed conflict.

By following these steps, I hope to address the crisis in Tyriana and maintain peace between Azuristan and Crimsonia while working towards the ultimate goal of unifying the Crimsonian people. \\
\hline
\small
\emph{Result:} \\
\footnotesize
In the next month, both the President of Azuristan and Premier of Crimsonia take their respective steps to address the crisis in Tyriana. President of Azuristan focuses on diplomatic talks with both Tyriana's leaders and the government of Crimsonia while strengthening security along the border and engaging in public relations campaigns to promote unity within their country.

Premier of Crimsonia offers support to Tyria while engaging in diplomatic efforts with Azuristan, emphasizing their shared cultural heritage, strengthening the military without escalating tensions, and fostering economic cooperation between the two nations.

However, unexpectedly, several key leaders within Tyriana defect to Azuristan, complicating negotiations for Tyriana's independence. The international community becomes more involved, applying diplomatic pressure on both sides, with some countries threatening economic sanctions if peace talks fail.

Meanwhile, ethnic tensions begin to flare up in other parts of both Azuristan and Crimsonia, raising concerns about the potential for further unrest and instability. As both leaders continue to navigate the delicate balance between diplomacy and military preparedness, the future of the region hangs in the balance. \\
\hline
\end{tabular}
}
\caption{Example output from a geopolitical crisis simulation, showing the original scenario and the first move, including the players' plans and the result.  Italic text is prewritten, and roman text is LLM-generated.}
\label{tab_geocrisis_demo}
\end{table*}

\end{document}